\documentclass[journal]{IEEEtran}

\usepackage{cite}
\usepackage{amsmath}
\usepackage{amssymb}
\usepackage{bm}
\usepackage{amsthm}

\usepackage{booktabs}
\usepackage{multirow}
\usepackage{tabularx}
\usepackage{float}
\usepackage{caption}
\usepackage{capt-of}
\usepackage{svg}
\usepackage{multirow}
\usepackage{makecell}
\usepackage{cuted}

\usepackage{algorithm}
\usepackage{algpseudocode}

\usepackage{xcolor}
\usepackage{soul}
\soulregister\cite7
\soulregister\ref7
\soulregister\pageref7

\usepackage{hyperref}
\hypersetup{colorlinks=true, linkcolor=black, citecolor=black, urlcolor=black}

\title{FU-MPC: Frontier- and Uncertainty-Aware Model Predictive Control
for Efficient and Accurate UAV Exploration with Motorized LiDAR}

\author{Jianping~Li, Pengfei~Wan, 
Zhongyuan~Liu,
Yi~Wang,
Yiheng~Chen,
Xinhang~Xu,
Rui~Jin,
Boyu~Zhou,
and~Lihua~Xie,~\IEEEmembership{Fellow, IEEE}%
\vspace{-1cm}
\thanks{This work was supported by NTUitive Gap Fund (NGF-2025-17006). J.~Li and P.~Wan contribute equally to this work.
J.~Li is the corresponding author.
(e-mail: \texttt{jianping.li@ntu.edu.sg})}}

\begin{document}
\maketitle
\markboth{IEEE/ASME TRANSACTIONS ON MECHATRONICS,~VOL.~XX, NO.~X, MONTH~YEAR}%
{Wan \MakeLowercase{\textit{et~al.}}: FU-MPC for UAV Exploration with Motorized LiDAR}



\begin{abstract}
Efficient UAV exploration in unknown environments requires rapid coverage expansion while maintaining accurate and reliable localization, since safe navigation in complex scenes depends on consistent mapping and pose estimation. However, for conventional LiDAR-equipped UAVs, the observable region is tightly coupled with the UAV pose and motion. Expanding coverage often requires additional translational or rotational maneuvers, which can reduce exploration efficiency and increase the risk of localization degradation in geometrically challenging environments. Motorized rotating LiDARs provide a promising solution by actively adjusting the sensor viewing direction without changing the UAV motion, thereby introducing an additional sensing degree of freedom. Nevertheless, existing exploration systems rarely exploit this scanning freedom as an explicit decision variable linked to both exploration progress and localization quality. To address this gap, we develop a UAV platform equipped with an independently actuated rotating LiDAR and propose a hierarchical exploration framework. The global planner organizes frontiers into representative viewpoints and sequences them using topology-aware transition costs. Built upon this planner, FU-MPC serves as a local receding-horizon scan controller that optimizes LiDAR rotation along the predicted flight trajectory. The controller jointly considers frontier-aware exploration utility and direction-dependent localization uncertainty, while lightweight surrogate evaluation enables real-time onboard execution. Experiments in complex environments demonstrate that the proposed system improves exploration efficiency while maintaining robust localization performance compared with fixed-pattern scanning and uncertainty-only baselines. The project page can be found at \url{https://kafeiyin00.github.io/FU-MPC/}.
\end{abstract}

\section{Introduction}

Autonomous UAV exploration in unknown environments is a key capability for rapid 3D mapping in applications such as underground infrastructure surveying and post-disaster assessment \cite{faria2019autonomous,zhang2023unmanned,kerle2019uav}. In these missions, the system must be both efficient, by rapidly expanding spatial coverage \cite{yamauchi1997frontier}, and accurate, by maintaining reliable localization and map consistency during long-horizon operation \cite{placed2023survey}. However, achieving both objectives remains challenging in complex environments with unevenly distributed geometric structures. Such conditions can degrade observability and cause SLAM errors to accumulate, which may compromise map consistency, destabilize autonomous navigation, and ultimately lead to mission failure \cite{xu2021fast,zhang2019beyond}.

\begin{figure}[]
    \centering
    \includegraphics[width=\linewidth]{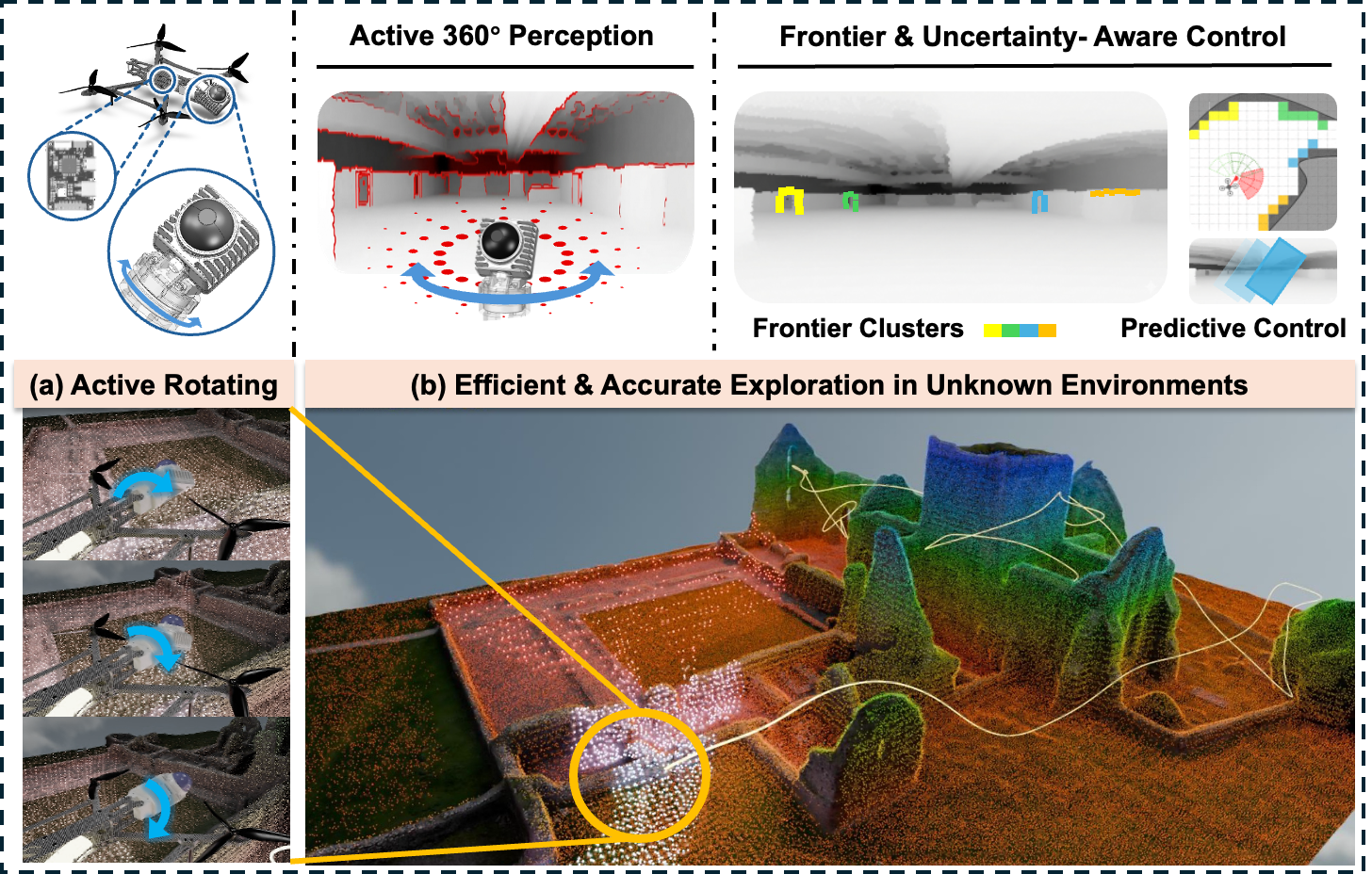}
    \caption{Key features of FU-MPC. (a) Mechanical design of the UAV with active rotating LiDAR; (b) Efficient and accurate exploration in unknown environments considering frontier distribution and positioning uncertainty.}
    \label{fig:abstract}
\end{figure}

One important reason for this difficulty is that the sensing capability of a UAV is tightly constrained by the LiDAR field of view (FoV). For rigidly mounted LiDARs, the observable region is directly coupled with the UAV pose and motion. As a result, expanding vertical or lateral coverage often requires additional translational or rotational maneuvers rather than only adjusting the sensing direction \cite{xu2022fast,chen2023self}. Such motion is inefficient in confined spaces and may further introduce LiDAR matching degeneracy, thereby increasing the burden on LiDAR-Inertial Odometry (LIO) \cite{chen2022direct,xu2021fast}. Therefore, limited and motion-coupled sensing becomes a key factor that restricts both exploration efficiency and localization robustness.
Motorized rotating LiDAR modules offer a promising alternative \cite{chen2022r,chen2023self}. By actively changing the scanning direction, they can enlarge the effective FoV without adding extra sensors or requiring excessive UAV maneuvers. More importantly, they introduce an additional actuation channel, namely the scan motion itself. This creates an opportunity to adjust the sensing process according to the current exploration and localization needs. However, although motorized LiDAR hardware has been increasingly studied, control algorithms that explicitly exploit this sensing degree of freedom for autonomous exploration remain limited.
Existing exploration systems still primarily optimize \emph{where} the UAV should move, while treating the LiDAR scanning mechanism as a preset peripheral process. Consequently, scan motion is rarely linked to the two task-level objectives that are critical during exploration: revealing frontier-rich unknown space and maintaining strong geometric constraints for localization. Active SLAM and information-theoretic planning methods address uncertainty reduction \cite{placed2023survey,placed2022general}, but they remain largely trajectory-centric and do not provide a direct, actionable mapping from controllable scan motion to exploration progress and estimation quality. What is still missing is a task-aware control formulation for the rotating LiDAR itself, so that the system can explicitly regulate scan motion according to exploration demand and localization condition. Without such a formulation, scan behavior remains trapped in a practical trade-off: faster sweeping can expose unknown space more quickly, whereas slower and more selective scanning can provide more stable geometric feature tracking for localization \cite{placed2023survey,chen2022direct}.

To address this gap, we design and build a UAV platform with an independently actuated motorized LiDAR module and tight multi-sensor time synchronization. Unlike existing platforms in which LiDAR rotation is treated as a passive or fixed-speed process, our system elevates LiDAR rotation speed to a dedicated real-time-commandable control input. A microcontroller-based timing hub aligns encoder, IMU, and LiDAR data streams under a unified timestamp protocol, enabling consistent sensor fusion across modalities. This hardware design decouples scanning motion from UAV flight dynamics and provides the physical foundation for treating scan motion as a decision variable in autonomous exploration.



Building on this platform, we redesign the global exploration planning pipeline for UAVs equipped with rotating LiDAR. Instead of sequencing frontier targets solely according to geometric distance, the proposed planner first aggregates frontier regions into representative viewpoints and then evaluates transitions between them using a topology-aware global cost. This cost incorporates path reachability, vertical motion burden, and heading consistency, enabling the planner to generate more executable global decisions in cluttered 3D environments. The resulting viewpoint sequence is subsequently passed to the local planner to produce dynamically feasible reference trajectories for downstream scan control \cite{zhou2021fuel}. Based on the planned flight trajectory, we further develop FU-MPC as a local receding-horizon scan controller to regulate the LiDAR rotation online. FU-MPC jointly optimizes frontier-aware exploration utility and direction-dependent localization quality, where the latter is estimated from local geometric constraints using a Fisher-information-based surrogate and an A-optimality criterion \cite{placed2023survey,zhang2019beyond}. To meet onboard real-time requirements, lightweight piecewise-linear surrogates are employed to approximate the predictive utility field and avoid expensive online evaluation \cite{tang2023bubble}. This design enables closed-loop scan optimization within a strict $0.1\,\text{s}$ control cycle. In summary, this paper makes the following contributions:

1) To address the limited exploration efficiency of existing UAV platforms that mostly rely on fixed sensing configurations, we develop a UAV with an independently actuated motorized LiDAR, enabling active scan motion as a real-time control input for exploration.

2) Existing global exploration planners mainly optimize the waypoint travelling distance. This often causes aggressive reorientation and inefficient exploration, and does not exploit the sensing characteristics of rotating-LiDAR UAVs. To address this issue, we propose a hierarchical planning framework that organizes frontiers into representative viewpoints and sequences them using a topology-aware global cost with path reachability, and heading consistency.

3) Safe exploration requires accurate localization, yet existing methods lack a local scan controller that jointly considers exploration efficiency and localization quality. To address this issue, we propose FU-MPC, which optimizes LiDAR scan direction along the predicted flight trajectory using frontier-aware utility and uncertainty metrics under real-time constraints.

4) We validate the proposed system through simulation in three large-scale scenes and real-world flight experiments, demonstrating improved exploration efficiency and accurate, robust SLAM performance compared with fixed-pattern scanning and uncertainty-only baselines.

\section{Related Work}

\subsection{Motorized LiDAR Platforms and Hardware Design}

Motorized rotating LiDARs can enlarge a UAV's effective field of view without using multiple sensors, but their integration introduces challenges in calibration, synchronization, and motion compensation. Since LiDAR measurements are affected by both UAV motion and actuator-induced scan motion, accurate modeling of the LiDAR--actuator transform and motion distortion is essential. Prior studies have addressed these issues through automatic self-calibration \cite{alismail2015automatic}, continuous-time trajectory modeling \cite{kaul2016continuous}, and real-time LiDAR odometry and mapping pipelines such as LOAM \cite{zhang2014loam}.

Recent hardware designs further improve the utility of actuated sensing through perception-feedback scan accumulation \cite{shi2023real}, variable-resolution panoramic scanning \cite{cui2024alphalidar}, lightweight robot-sensor decoupling \cite{chen2024design}, and self-rotating UAV platforms \cite{chen2023self}. However, most existing systems still treat LiDAR rotation as a preset peripheral process rather than a task-driven control input. In contrast, our platform provides a tightly time-synchronized and real-time-commandable scan-speed interface, enabling LiDAR rotation to be actively regulated according to exploration progress and localization quality.

\subsection{Frontier-Based Exploration and Viewpoint Planning}

A large body of UAV exploration research focuses on \emph{where the robot should move}. Offline methods pre-compute coverage paths from coarse geometry or aerial imagery \cite{feng2024fc,yan2021sampling,zhou2020offsite,feng2023predrecon}, but they are difficult to apply when accurate prior maps are unavailable. Online frontier-based methods instead guide UAVs toward boundaries between known and unknown space \cite{yamauchi1997frontier,yamauchi1998frontier,zhou2021fuel,zhou2023racer,cao2021tare,tang2023bubble}. Recent studies further incorporate execution cost, dynamic feasibility, and global routing objectives, such as TSP-style topology planning, to reduce redundant revisits and improve exploration efficiency \cite{bircher2016receding,lindqvist2024tree,cieslewski2017rapid,zhou2021fuel,zhou2023racer}. Gimbal-based or stabilized-payload systems decouple sensor pointing from UAV heading to reduce unnecessary yawing \cite{xu2024vrexplorer}, but their vertical sensing flexibility remains limited.

Despite these advances, most methods still treat LiDAR sensing as a fixed pattern: the UAV trajectory is optimized, while the scanner operates at a preset rate independent of local exploration and localization needs. Consequently, scan effort cannot be actively directed toward frontier-rich or geometrically informative regions, nor reduced when localization is vulnerable in feature-sparse areas. In contrast, our method couples predictive frontier gain with motorized scan control in a receding-horizon framework, enabling scan effort to be allocated according to both exploration value and localization quality.

\subsection{Active SLAM and Uncertainty-aware Scan Control}

Active SLAM links sensing, motion, and estimation by selecting actions that improve state inference quality. Observability-aware methods analyze which state components can be reliably estimated and preserve the correct observable subspace during linearization \cite{huang2010observability}. Information-theoretic planning further selects viewpoints or trajectories by maximizing expected information gain to reduce pose and map uncertainty \cite{burgard2005information}, while belief-space control explicitly propagates covariance dynamics for posterior uncertainty reduction \cite{koga2022active}. Learning-based active perception complements these methods by using semantic cues to improve observation reliability and informativeness \cite{bartolomei2021semantic}.

The most relevant work to ours is UA-MPC, which treats LiDAR spin rate as a control input and optimizes it online to balance sensing efficiency and localization accuracy through a predicted estimation utility \cite{li2025ua}. Related MPC and inverse-reinforcement-learning methods also provide useful tools for closed-loop decision-making under constraints \cite{romero2024actor,cao2023trust}.

However, most existing active SLAM methods remain trajectory-centric and often rely on costly belief propagation or matrix operations, making them difficult to deploy for high-rate onboard scan control. More importantly, they do not directly map controllable scan motion to direction-dependent localization quality, nor do they couple estimation-aware scan control with frontier-driven exploration. To address this gap, we construct a Fisher-information-based uncertainty surrogate tailored to motorized scan actuation, integrate it with predictive frontier gain in a unified MPC objective, and enable real-time execution through lightweight piecewise-linear approximations.

\section{System Overview}

\subsection{Hardware System for the Motorized Scanning UAV}



\begin{figure}[]
    \centering
    \includegraphics[width=\columnwidth]{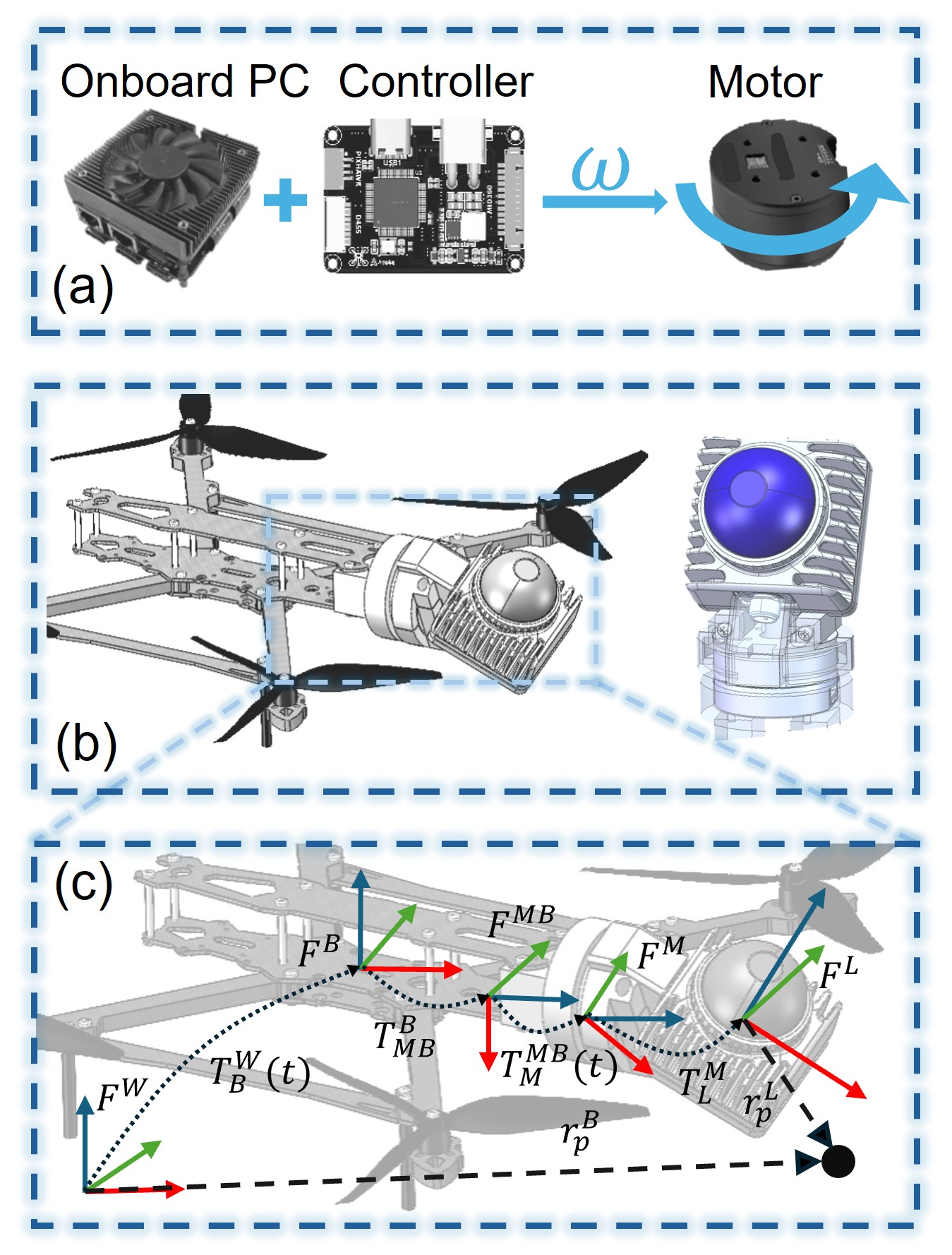}
    \caption{ Hardware system design of the active rotating LiDAR-UAV. (a) LiDAR rotating control system. (b) UAV with active rotating LiDAR hardware design. (c) Coordination systems related to the FU-MPC UAV. }
    \label{fig:hardware}
\end{figure}

To enable active exploration, we developed a tightly time-synchronized UAV platform (Fig.~\ref{fig:hardware}). An STM32 microcontroller hardware-synchronizes multi-sensor streams (LiDAR, IMU, encoder, and flight states) and transmits them to the onboard computer, which computes real-time motor commands. A front-mounted LiDAR rotates independently of UAV dynamics, providing panoramic perception. This variable-speed actuation allows the system to adaptively prioritize exploration frontiers and regions with high localization uncertainty without compromising flight stability.

\subsection{Notation}

We define four coordinate frames (Fig.~\ref{fig:hardware}): world $\mathcal{F}^{W}$, UAV body $\mathcal{F}^{B}$, motor base $\mathcal{F}^{MB}$, and LiDAR $\mathcal{F}^{L}$. A point $\mathbf{r}_p^L$ maps to the world frame $\mathbf{r}_p^W$ via the kinematic chain:
\begin{equation}
\begin{aligned}
\mathbf{r}_p^W
&= \mathbf{R}_B^W(t)\Big(
\mathbf{R}_{MB}^B \big(
\mathbf{R}_{M}^{MB}(\theta(t))
\big( \mathbf{R}_L^M \mathbf{r}_p^L + \mathbf{r}_L^M \big)
\big)
+ \mathbf{r}_{MB}^B
\Big) \\
&\quad + \mathbf{r}_B^W(t),
\end{aligned}
\end{equation}
where $(\mathbf{R}_L^M, \mathbf{r}_L^M)$ and $(\mathbf{R}_{MB}^B, \mathbf{r}_{MB}^B)$ are constant extrinsics for the LiDAR and motor base, respectively. The LIO estimates the UAV pose $(\mathbf{R}_B^W(t), \mathbf{r}_B^W(t))$.

The rotating motor stage introduces a time-varying rotation $\theta(t)$ about the $z$-axis of $\mathcal{F}^{MB}$:
\begin{equation}
\mathbf{R}_{M}^{MB}(\theta(t)) = \begin{bmatrix}
\cos\theta(t) & -\sin\theta(t) & 0 \\
\sin\theta(t) &  \cos\theta(t) & 0 \\
0             &  0             & 1
\end{bmatrix}.
\end{equation}

For FU-MPC, time is discretized by a fixed interval $\Delta t$. The predicted motor state evolves as:
\begin{equation}
\theta_{i+1} = \theta_i + \omega_i \Delta t,
\end{equation}
where the commanded angular velocity $\omega_i$ is the optimization variable.

\subsection{Hierarchical Exploration Planning}

We build the overall exploration pipeline upon the point-cloud-based framework~\cite{geng2025epic} and adapt it to a UAV equipped with a motorized rotating LiDAR. Section \ref{sec:global} focuses on global planning: it converts raw frontier observations into representative viewpoints, constructs a topology-aware global sequencing strategy, and generates dynamically feasible local reference trajectories using MINCO. Built on top of this motion-planning layer, Section \ref{sec:local} focuses on active perception: given the future flight trajectory produced by Section \ref{sec:global}, FU-MPC optimizes the LiDAR rotation online to improve frontier discovery and localization quality. In this way, Section \ref{sec:global} determines where the UAV should move next, whereas Section \ref{sec:local} determines how the motorized LiDAR should scan along that motion.

\begin{figure}[]
    \centering
    \includegraphics[width=\columnwidth]{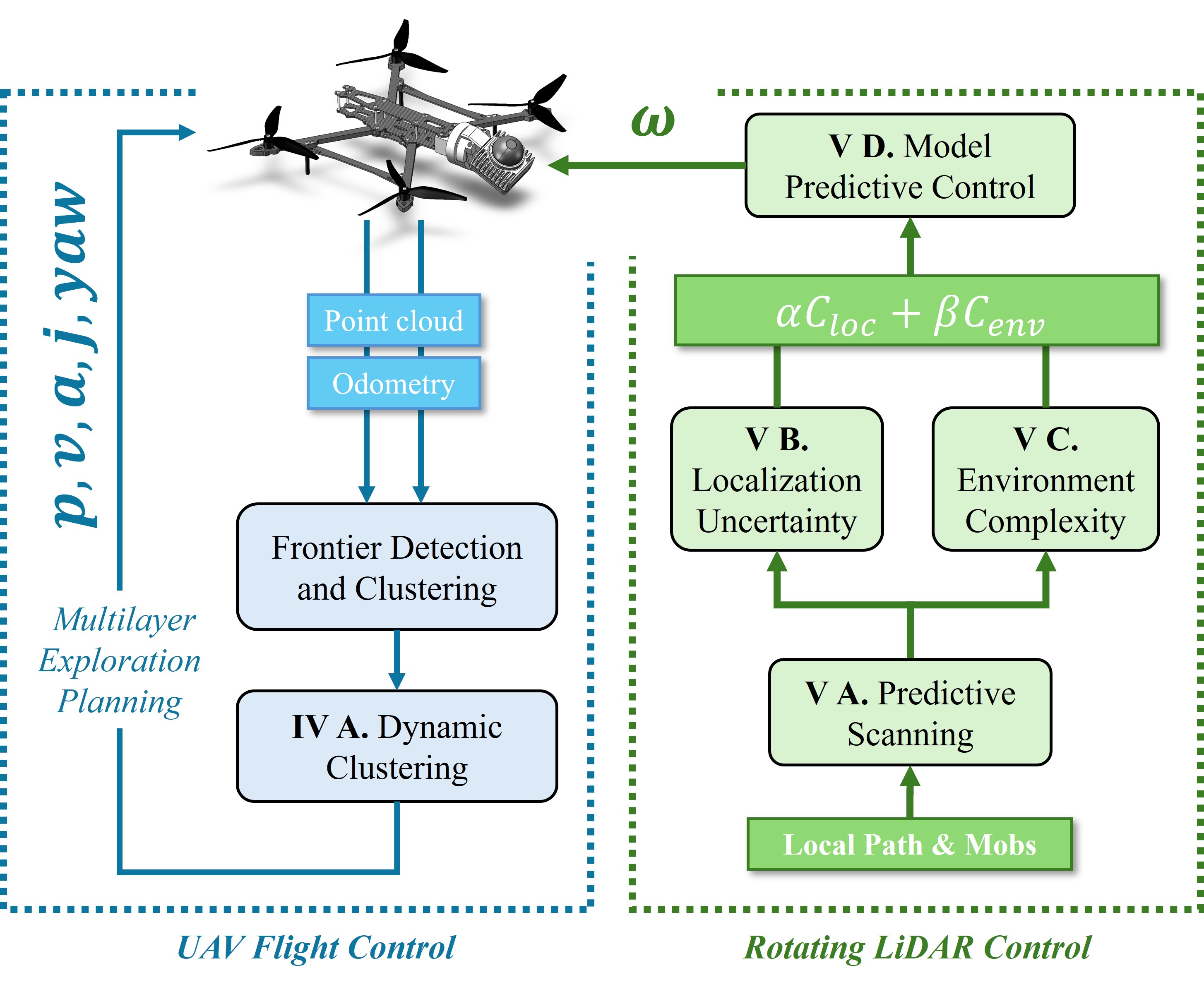}
    \caption{Overview of the proposed FU-MPC system for exploration. The system integrates hierarchical exploration planning, predictive frontier gain estimation, direction-dependent localization uncertainty evaluation, and surrogate-assisted receding-horizon optimization to adapt the motorized LiDAR rotation speed online for efficient exploration and reliable SLAM.}
    \label{fig:wide}
\end{figure}

\section{Topology-Aware Global Exploration Planning}\label{sec:global}

\begin{figure*}[]
    \centering
    \includegraphics[width=.9\textwidth]{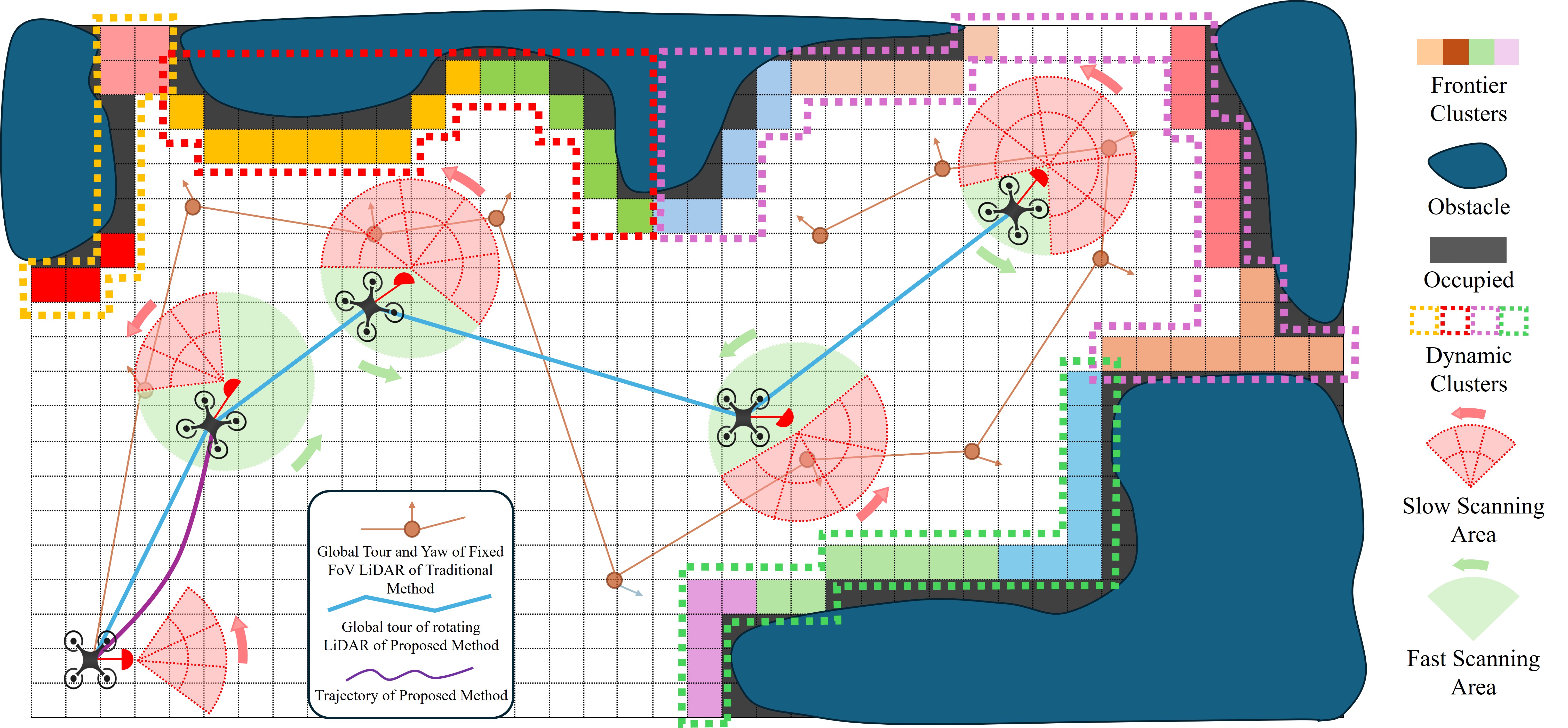}
    \caption{The process of global planning and the global tour compared to the fixed FoV LiDAR method.}
    \label{fig:global_planning}
\end{figure*}

\subsection{Viewpoint Sampling and Dynamic Clustering}

To reduce target redundancy while preserving observation structure, we organize frontier clusters into a higher-level representation. For each frontier cluster \(FC\), we first estimate its center \(c\) and dominant normal \(n\), and then sample candidate viewpoints only in the local hemisphere oriented by \(n\). This design is motivated by the fact that informative observations are usually obtained from the free-space side of a frontier, whereas back-side or tangential views provide limited exploration gain.

Each sampled viewpoint is required to satisfy three conditions: it must lie inside the valid map boundary, keep a safe distance from nearby obstacles, and belong to a region supported by the current topological graph. For every remaining candidate viewpoint \(v\), we evaluate its visibility with respect to the frontier cluster by voxel ray-casting. If the cluster contains \(N_{FC}\) frontier cells, its visibility ratio is defined as:
\begin{equation}
r(v)=\frac{1}{N_{FC}}\sum_{k=1}^{N_{FC}}\delta(v,p_k),
\label{eq:viewpoint_visibility}
\end{equation}
where \(p_k\) is the \(k\)-th frontier cell and \(\delta(v,p_k)\in\{0,1\}\) indicates whether \(p_k\) is visible from \(v\). A viewpoint is regarded as valid if \(r(v)\) exceeds a preset threshold \(r_0\). Accordingly, each frontier cluster is associated with an effective visible region
\begin{equation}
R=\{\,v \mid r(v)\ge r_0\,\}.
\label{eq:visible_region}
\end{equation}

Based on these visible regions, we further build dynamic clusters. The key idea is that if multiple frontier clusters share overlapping visible regions, then they can be jointly observed and need not be treated as independent planning targets\cite{zhu2025flare}. Let \(R_s\) denote the shared visible region of an existing dynamic cluster and \(R_d\) denote the  dynamic cluster after merging. A new frontier cluster with visible region \(R\) can be merged only if \(R\cap R_d \neq \varnothing\). To balance shared observability and spatial compactness, we define the merge cost as
\begin{equation}
J=\alpha \Delta d-\beta q,
\label{eq:merge_cost_short}
\end{equation}
where \(\Delta d\) is the increase in the spatial extent after merging, \(q\) is the size of the shared visible region, and \(\alpha,\beta\) are weighting coefficients. We greedily merge each frontier cluster into the dynamic cluster with the minimum valid merge cost; otherwise, a new dynamic cluster is created. In Fig.~\ref{fig:global_planning}, the frontier clusters are detected through the observation map and after viewpoint sampling, the dynamic clusters will be constructed by the method above.

Finally, a representative viewpoint is selected for each dynamic cluster from its shared visible region. In this way, the original frontier set is transformed into a much smaller set of representative observation targets, which preserves the joint visibility structure while significantly reducing the complexity of the subsequent global tour planning. In Fig.~\ref{fig:global_planning},  the proposed viewpoint sampling and dynamic clustering strategy reduces redundant frontier targets and produces a more compact global tour than the fixed-FoV baseline. This demonstrates that exploiting the enlarged observation capability of the rotating LiDAR can improve the efficiency of global exploration planning.

\subsection{Global Waypoint Sequencing with Topology-Aware Cost}

After dynamic cluster construction, the planner obtains a set of representative viewpoints \(\mathcal{V}=\{v_1,v_2,\dots,v_M\}\). The objective of the global planning layer is to determine an efficient visiting order over these high-level observation targets and provide the current \emph{next goal} for the local planner. The key idea is that, for exploration in cluttered 3D environments, the straight-line distance between viewpoints is often a poor proxy for the actual execution effort. We therefore design a topology-aware global cost that explicitly captures the traversability and motion burden of moving between representative viewpoints, which is one of the main contributions of this planning module.

Specifically, each representative viewpoint is temporarily inserted into the topological graph, and the current odometry node is treated as the start node. For any two nodes \(i\) and \(j\), if a feasible topological path exists, their pairwise transition cost is defined as
\begin{equation}
d_{ij}=\frac{1}{v_{\max}/2}\sum_{l}
\Bigl(\|x_{l+1}-x_l\|+0.5|z_{l+1}-z_l|\Bigr),
\label{eq:tour_cost_short}
\end{equation}
where \(x_l\) is the \(l\)-th point on the path, \(z_l\) is its altitude, and \(v_{\max}\) is the maximum flight speed. Unlike a Euclidean metric, this cost is evaluated along a feasible topological route and additionally penalizes altitude variation, so that the resulting value reflects not only path length but also the vertical maneuvering effort required by the UAV. In this way, the global cost favors transitions that are shorter, smoother, and easier to execute, while unreachable pairs are assigned a large penalty to exclude infeasible connections from the global tour. The generated shorter, smoother, and easier global tour is shown in Fig.~\ref{fig:global_planning}.

In addition, the first selected target has a direct influence on the immediate motion of the UAV and thus strongly affects short-horizon execution efficiency. To reduce abrupt reorientation and improve consistency between the global plan and the current flight state, we introduce a heading-consistency term for the first target:
\begin{equation}
d^{\text{head}}_{j}=w_f{\theta_j},
\label{eq:heading_penalty_short}
\end{equation}
where \(\theta_j\) is the angle between the current velocity direction and the direction of \(v_j\), and \(w_f\) is a weighting coefficient. This term biases the planner toward targets that are better aligned with the current motion direction, thereby reducing unnecessary turning at the beginning of each replanning cycle.

With all reachable viewpoints, the pairwise transition costs are assembled into a global cost matrix, and the sequencing problem is formulated as an open tour optimization:
\begin{equation}
J_{\text{tour}}=\sum_{k=0}^{M-1} d_{\pi_k,\pi_{k+1}},
\label{eq:global_tour_short}
\end{equation}
where \(\pi=(0,\pi_1,\pi_2,\dots,\pi_M)\) is the visiting order and node \(0\) denotes the current odometry node. Since exploration does not require returning to the start node, the problem is solved as an open asymmetric traveling salesman problem (ATSP) using the Lin--Kernighan--Helsgaun (LKH) solver. Overall, this global planning design converts frontier-aware representative viewpoints into an execution-oriented visiting sequence through a cost formulation that jointly accounts for topological reachability, vertical transition burden, and motion-direction consistency.

\section{FU-MPC for Local Scan Control}\label{sec:local}

During UAV exploration, the rotating LiDAR continuously scans the surrounding environment along the planned flight trajectory. Its angular velocity directly affects both the distribution of newly observed frontiers and the geometric constraints available for LiDAR-inertial localization. Therefore, we formulate LiDAR rotation regulation as a receding-horizon control problem, in which the scan motion is optimized online according to predicted exploration gain and localization uncertainty.

\subsection{Predictive Scanning}
Traditional rotating LiDAR typically operates at a fixed rotational speed, where its scanning behavior is determined entirely by the current motor state and cannot utilize the future motion information of the UAV. However, in active exploration tasks, the effective observation within a future period depends not only on the current orientation of the LiDAR but also on where the UAV will perform the scanning. Therefore, this paper predicts the future scanning process of the LiDAR within the current control horizon, which is shown in Fig~\ref{fig:abstract}, and uses this prediction result as the foundation for the subsequent localization uncertainty evaluation and environmental complexity assessment.

Specifically, at the current control time $t_0$, the system knows the current angle $\theta_0$ and angular velocity $\omega_0$ of the rotating LiDAR, and obtains the future reference trajectory $\mathbf{p}_{\mathrm{ref}}(t)$ of the UAV from the local planner. Over a prediction horizon of length $N$, we represent the scanning state at the $i$-th prediction step as:
\begin{equation}
\mathbf{x}_i=
\begin{bmatrix}
\theta_i\\
\omega_i
\end{bmatrix},
\label{eq:scan_state}
\end{equation}
where $\theta_i$ and $\omega_i$ represent the LiDAR angle and angular velocity at that moment, respectively. The control input is defined as the angular acceleration $u_i$. Considering the velocity upper bounds and angular acceleration constraints of the motor, the scanning dynamics adopt the following discrete form:
\begin{equation}
\omega_{i+1}=\mathrm{sat}_{[\omega_{\min},\,\omega_{\max}]}\!\left(\omega_i+u_i\Delta t\right),
\label{eq:scan_omega}
\end{equation}
\begin{equation}
\theta_{i+1}=\theta_i+\omega_{i+1}\Delta t,
\label{eq:scan_theta} 
\end{equation}
where $\Delta t$ is the prediction step size, and $\mathrm{sat}[\cdot]$ represents the saturation operator. Thus, given the initial state $(\theta_0,\omega_0)$ and the control sequence $\{u_i\}_{i=0}^{N-1}$, the future scanning state sequence within the current control horizon can be obtained.

To couple the scanning prediction with the UAV motion, we further use the future reference trajectory to obtain the sensor position at each prediction step:
\begin{equation}
\mathbf{p}_i=\mathbf{p}_{\mathrm{ref}}(t_0+i\Delta t).
\label{eq:sensor_pos}
\end{equation}
Combining the attitude transformation from the body frame to the world frame $\mathbf{R}_{B}^{W}$ and the fixed extrinsic parameters from the LiDAR motor base frame to the body frame $\mathbf{R}_{MB}^{B}$, the principal scanning direction of the LiDAR in the world frame at the $i$-th prediction step can be written as:
\begin{equation}
\mathbf{d}_i^W=\mathbf{R}_{B}^{W}\mathbf{R}_{MB}^{B}
\begin{bmatrix}
\cos\theta_i\\
\sin\theta_i\\
0
\end{bmatrix}.
\label{eq:scan_direction}
\end{equation}
Therefore, at each prediction step, the system can obtain a set of future scanning states $(\mathbf{p}_i,\theta_i,\omega_i,\mathbf{d}_i^W)$. This implies that what this paper optimizes is not the instantaneous scanning angle at a certain moment, but a continuous scanning process evolving along the future flight trajectory.

Based on the above prediction results, the future observation area within the current control horizon can be further determined, and the impact of different scanning strategies on localization constraints and environmental perception gains can be analyzed accordingly. In other words, the role of Predictive Scanning is not to directly evaluate the quality of the scan, but to provide a unified time-domain prediction basis for the subsequent two analysis modules. The former evaluates the constraint capability of the local geometry on pose estimation based on the predicted scanning states, while the latter evaluates the coverage gain of future observations on the frontier areas based on the same predicted scanning states.

\subsection{Localization Uncertainty Analysis}

Reliable localization requires the UAV to observe surfaces under sufficiently diverse geometric constraints. To quantify the localization quality associated with different scan directions, we construct a Fisher-information-based metric from the distribution of visible surface normals. 
For each retained point, the surface normal $\mathbf{n}_k$ is estimated from its $K$ nearest neighbors. For a candidate motor angle $\theta_c$, the set of normals visible from that direction defines the Fisher information matrix:
\begin{equation}
  \mathbf{F}(\theta_c) = \sum_{k} \mathbf{n}_k \mathbf{n}_k^\top
  \in \mathbb{R}^{3\times 3}.
\end{equation}
A well-conditioned matrix $\mathbf{F}$ indicates that the observed local geometry provides balanced constraints along different directions. Based on this matrix, we define the localization cost as the trace of the inverse of the Fisher information matrix,
\begin{equation}
  f(\theta_c)
  = \mathrm{tr}\!\left(\mathbf{F}(\theta_c)^{-1}\right),
\end{equation}
A smaller value of $f(\theta_c)$ indicates stronger expected localizability for the corresponding scan direction. To satisfy real-time control requirements, we evaluate this cost offline within each update cycle at $N_c$ uniformly spaced candidate angles, and store the results in a lookup table. During MPC, the localization cost at an arbitrary angle is obtained by linear interpolation between neighboring samples. This design separates the expensive normal-estimation and matrix-evaluation steps from the high-rate optimization loop while preserving direction-dependent uncertainty information for online control.

\subsection{Frontier-Aware Exploration Utility}

To explicitly connect scan motion with exploration progress, we construct a frontier-aware environment complexity cost that evaluates how much frontier information can be revealed by a predicted scan pose. For each frontier cluster with center $\mathbf{c}_i$, visibility is determined jointly by a distance gate and an angular gate relative to the current scan direction. Visible clusters contribute a reward weighted by both spatial proximity and cluster intensity:
\begin{equation}
  r_{k,i} = \exp\!\left(-\frac{d_{k,i}^2}{2\sigma^2}\right) \cdot s_i,
\end{equation}
where $d_{k,i}=\|\mathbf{p}_k-\mathbf{c}_i\|$, $\sigma$ is a distance-decay scale parameter and $s_i$ denotes the cluster intensity score. Based on these visible rewards, the complexity cost at prediction step $k$ is defined as
\begin{equation}
  J(\theta_k)
  = R_{\mathrm{total}} - \sum_{i \in \mathcal{F}(\theta_k)} r_{k,i},
\end{equation}
where $R_{\mathrm{total}}=\sum_{i} r_{k,i}$ is a constant baseline at prediction step k and $\mathcal{F}(\theta_k)$ is the set of frontier clusters visible at angle $\theta_k$. A smaller value of $J(\theta_k)$ therefore indicates that the candidate scan is expected to expose more informative frontiers. This design provides a lightweight surrogate for frontier gain and establishes a direct optimization link between motorized scanning and exploration efficiency.

\subsection{Frontier-Uncertainty Model Predictive Control}

FU-MPC unifies the frontier-aware complexity term and the uncertainty-aware localization term into a single receding-horizon controller that regulates the LiDAR motor online. Over the prediction horizon, the controller minimizes:
\begin{equation}
  \min_{\{u_k\}} \;
  \sum_{k=0}^{N-1}
  \Bigl[
    \alpha J_(\theta_k)
    + \beta f(\theta_k)
  \Bigr]
  + \gamma \sum_{k=0}^{N-2}(u_{k+1}-u_k)^2
  \label{eq:mpc_obj},
\end{equation}
subject to:
\begin{align}
  \omega_{\min} &\leq \omega_k \leq \omega_{\max}, \label{eq:vel_bound}\\
  |u_k| &\leq u_{\max}, \label{eq:acc_bound}
\end{align}
where the first two terms balance exploration gain and localization quality, and the third term regularizes control variation to avoid aggressive speed changes. $\alpha$, $\beta$ and $\gamma$ are the weights of different cost functions. N is the prediction window length.

To enable real-time onboard execution, the nonlinear objective in MPC is optimized using a local approximation strategy. At each MPC update, the cost terms are linearized around the current predicted scan trajectory, yielding a lightweight local problem solved by Ceres. The optimization follows a receding-horizon scheme, where only the first scan command is applied to the motor and the remaining sequence is shifted forward to warm-start the next iteration. The uncertainty table is refreshed asynchronously in a separate thread, thereby separating model updating from online scan optimization and maintaining low-latency operation under onboard computational constraints.

\section{Experiments}


In this section, we implemented FU-MPC using C++ and ROS. We evaluate our method against three state-of-the-art exploration algorithms: FUEL\cite{zhou2021fuel}, EPIC\cite{geng2025epic}, ERRT\cite{lindqvist2024tree}. These three methods show different ways of real-time environmental information processing. As state-of-the-art exploration techniques, they provide a fair and reproducible basis for comparison. 

\subsection{Baselines and Metrics}
\subsubsection{Baselines}

We  evaluate our method against three state-of-the-art exploration algorithms: EPIC, FUEL, ERRT. Our tests were conducted in three large-scale scenes: a sinkhole $[70 \times 45 \times 13.5] \, \mathrm{m}^3$, a spatial maze $[119 \times 20 \times 8] \, \mathrm{m}^3$, and a lava tube $[315 \times 53.5 \times 27] \, \mathrm{m}^3$. We rendered all three scenarios into point cloud maps. Each method was executed three times per scene with a maximum velocity of 3.0 m/s.

\subsubsection{Metrics}

To quantitatively evaluate the accuracy of localization and mapping via FU-MPC, we adopt the Absolute Pose Error (APE) as one of the evaluation metrics of FU-MPC. APE measures the deviation between the estimated UAV trajectory of exploration and the ground-truth poses, providing a direct indicator of localization and mapping accuracy. 

To measure exploration efficiency, we used the exploration coverage rate over the same time period length as an indicator in the simulation experiment.

\subsection{Simulation Evaluation}

\subsubsection{Efficiency of Exploration}

\begin{figure*}[]
    \centering
    \includegraphics[width=.9\textwidth]{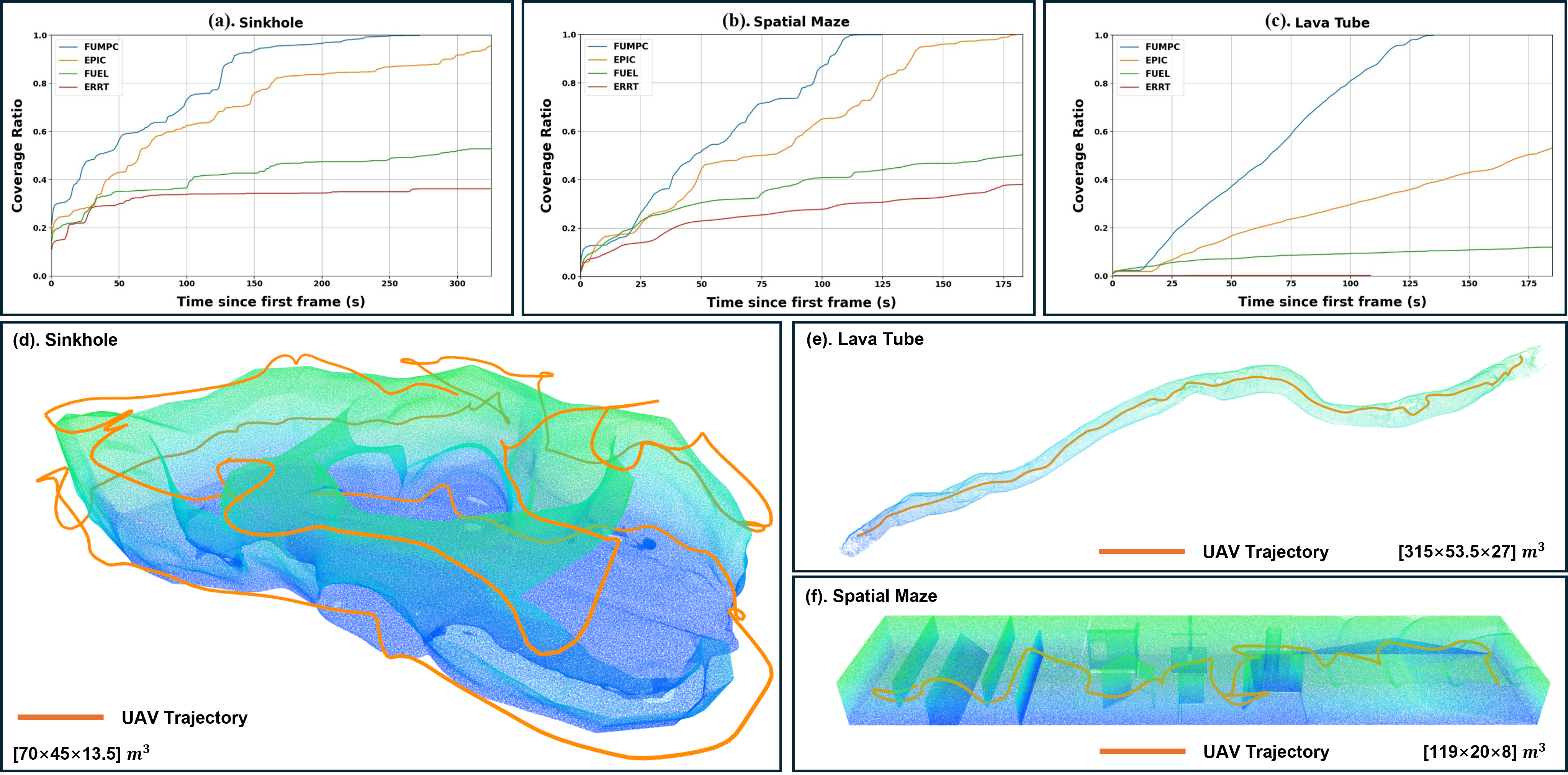}
    \caption{ The exploration progress of all three state-of-the-art benchmarks and the proposed method in three scenes. (a) The exploration progress of four methods in the Sinkhole. (b) The exploration progress of four methods in the Spatial Maze. (c) The exploration progress of four methods in the Lava Tube. (d)-(f) The UAV trajectories of the proposed FU-MPC in Sinkhole, Lava Tube, and  Spatial Maze.}
    \label{fig:Baseline_coverage}
\end{figure*}

We evaluate the exploration performance on three representative 3D maps with distinct geometric and perceptual challenges to demonstrate the benefits of active rotating LiDAR and the proposed FU-MPC planner. Table~\ref{tab:simulation_benchmark_results} reports the exploration time, trajectory length, and completion status, while Fig.~\ref{fig:Baseline_coverage} shows the coverage evolution over time. Across all scenarios, FU-MPC achieves faster coverage accumulation and shorter completion time than EPIC. Notably, only FU-MPC and EPIC complete the large-scale benchmarks, whereas FUEL and ERRT fail within the effective time budget, with their coverage curves saturating at low values.

\begin{table}[t]
    \centering
    \caption{Results of Benchmark Comparisons}
    \label{tab:simulation_benchmark_results}
    \renewcommand{\arraystretch}{1.2}
    
    \begin{tabular}{ccccccc}
        \hline \hline

        \textbf{Scene} & \textbf{Method} & \makecell{\textbf{Exp.} \\ \textbf{Time (s)}} & \textbf{Trajectory (m)} & \makecell{\textbf{Exp.} \\ \textbf{Fin.}} \\
        \hline
        
        \multirow{4}{*}{Spatial Maze} 
        & ERRT & -- & -- & No \\
        & FUEL & -- & -- & No \\
        & EPIC & 178.65 & 418.80 & \textbf{Yes} \\
        & FU-MPC  & \textbf{112.98} & \textbf{214.94} & \textbf{Yes} \\
        \hline
        
        \multirow{4}{*}{Sinkhole} 
        & ERRT  & -- & -- & No \\
        & FUEL  & -- & -- & No \\
        & EPIC  & 405.99 & 779.06 & \textbf{Yes} \\
        & FU-MPC  & \textbf{255.01} & \textbf{607.63} & \textbf{Yes} \\
        \hline
        
        \multirow{4}{*}{Lava tube} 
        & ERRT & -- & -- & No \\
        & FUEL & -- & -- & No \\
        & EPIC & 359.91 & 338.67 & \textbf{Yes} \\
        & FU-MPC  & \textbf{138.11} & \textbf{343.17} & \textbf{Yes} \\
        
        \hline \hline
    \end{tabular}
\end{table}

The Spatial Maze is a large multi-level environment with ramps, height-varying barriers, and vertically separated passages. In such 3D scenes, a front-mounted LiDAR may miss regions below the UAV due to limited vertical FoV, leading to incomplete exploration or navigation failures near obstacles. The active rotating LiDAR provides both upward and downward observations, improving traversability understanding and reducing unobserved areas. FU-MPC completes the mission in 112.98 s, which is 36.8\% faster than EPIC at 178.65 s. It also reduces the trajectory length from 418.80 m to 214.94 m, corresponding to a 48.7\% reduction and indicating fewer redundant revisits. The coverage curves show that FU-MPC reaches nearly full coverage around 110 s, whereas EPIC approaches full coverage much later. In contrast, FUEL and ERRT remain at low coverage and fail to complete this large-scale maze.

The Sinkhole represents an open, large-scale environment where global visibility and viewpoint selection dominate exploration efficiency. By expanding the sensing footprint and mitigating blind directions, rotating LiDAR improves frontier discovery and reduces unproductive revisits. As shown in Table~\ref{tab:simulation_benchmark_results}, FU-MPC finishes in 255.01 s, compared with 405.99 s for EPIC, achieving a 37.2\% reduction in exploration time. FU-MPC also shortens the trajectory from 779.06 m to 607.63 m, corresponding to a 22.0\% reduction. As shown in Fig.~\ref{fig:Baseline_coverage}, FU-MPC maintains higher coverage throughout the run and reaches near-complete coverage earlier, while EPIC grows more slowly, especially in the initial stage. FUEL and ERRT again plateau at low coverage, indicating limited scalability in large open scenes.

The Lava Tube is a long and narrow tunnel-like environment, relevant to future lunar subsurface exploration. Its corridor-constrained geometry limits route diversity, so successful methods tend to follow similar routes and thus produce comparable travel distances. This explains the close trajectory lengths of EPIC and FU-MPC in Table~\ref{tab:simulation_benchmark_results}, 338.67 m and 343.17 m, respectively, with only a 1.3\% difference. However, FU-MPC still achieves a substantial time reduction, completing in 138.11 s compared with 359.91 s for EPIC, corresponding to a 61.6\% improvement.This time gap is mainly attributed to the different sensing mechanisms. EPIC relies on UAV yaw rotation to adjust the sensing direction of the front-mounted LiDAR, which introduces frequent heading changes and prevents the UAV from accelerating continuously to $v_{max}$ along the lava tube. The coverage curves further confirm this trend: FU-MPC reaches maximum coverage roughly 50 s earlier than EPIC within the shown time window, demonstrating more consistent mapping progress in elongated, visibility-limited tunnels. Meanwhile, FUEL and ERRT make little progress and do not finish within the mission time budget.

Overall, FU-MPC reduces completion time by 36.8\%, 37.2\%, and 61.6\% in Spatial Maze, Sinkhole, and Lava Tube, respectively, compared with EPIC. It also reduces path length by 48.7\% and 22.0\% in the first two scenes. Together with the consistently steeper coverage growth in Fig.~\ref{fig:Baseline_coverage} and the failure of FUEL/ERRT to complete any benchmark, these results validate the efficiency and robustness of FU-MPC with active rotating LiDAR for complex 3D exploration tasks.

\begin{table}[t]
\centering
\caption{Comparison of FU-MPC with baseline methods on three simulation maps.}
\label{tab:baseline_comparison}
\begin{tabular}{cccccc}
\hline\hline
\textbf{Scene} & \textbf{Method} & \makecell{\textbf{Exp.} \\ \textbf{Fin.}} & \textbf{RMSE (m)} & \textbf{Mean (m)} & \textbf{Max (m)} \\
\hline 
\multirow{4}{*}{\makecell{Spatial \\ Maze}}
& EPIC & \textbf{Yes} & 0.112 & 0.098 & 0.545 \\
& ERRT & No & 0.019 & 0.014 & 0.057 \\
& FUEL & No & 0.025 & 0.020 & 0.095 \\
& FU-MPC & \textbf{Yes} & \textbf{0.068} & \textbf{0.063} & \textbf{0.161} \\
\hline
\multirow{4}{*}{\makecell{Lava \\ Tube}}
& EPIC & \textbf{Yes} & 0.664 & 0.542 & 1.821 \\
& ERRT & No & 0.118 & 0.104 & 0.303 \\
& FUEL & No & 0.076 & 0.067 & 0.270 \\
& FU-MPC & \textbf{Yes} & \textbf{0.191} & \textbf{0.153} & \textbf{1.418} \\
\hline
\multirow{4}{*}{Sinkhole}
& EPIC & \textbf{Yes} & 58.987 & 22.502 & 557.526 \\
& ERRT & No & 0.021 & 0.019 & 0.116 \\
& FUEL & No & 0.072 & 0.048 & 0.813 \\
& FU-MPC & \textbf{Yes} & \textbf{1.030} & \textbf{0.894} & \textbf{2.426} \\
\hline\hline
\end{tabular}
\end{table}

\subsubsection{Accuracy of Mapping}

Table~\ref{tab:baseline_comparison} compares FU-MPC with EPIC, FUEL, and ERRT in terms of task completion status and translational APE.
It should be noted that APE is computed only along the executed trajectory.
Therefore, a low APE does not necessarily indicate successful exploration, especially when the planner falls into local repetitive scanning.

In our experiments, FUEL and ERRT failed to complete the exploration task in several environments.They tended to repeatedly scan already explored regions instead of continuously expanding toward unknown space.This local repetitive behavior resulted in relatively small pose variations and geometrically familiar observations, which explains their low APE values.
Therefore, their low translational APE should be interpreted together with their incomplete exploration status.

Compared with EPIC, FU-MPC achieves lower translational APE while successfully completing the exploration task.In the Spatial Maze map, FU-MPC reduces the RMSE from 0.112~m to 0.068~m and the maximum APE from 0.545~m to 0.161~m.In the Lava Tube map, FU-MPC reduces the RMSE from 0.664~m to 0.191~m and the maximum APE from 1.821~m to 1.418~m.This indicates that the proposed method improves localization accuracy over EPIC while maintaining successful exploration.

In the Sinkhole map, EPIC suffers from severe trajectory drift, leading to an RMSE of 58.987~m and a maximum APE of 557.526~m.In contrast, FU-MPC successfully completes the exploration task and maintains bounded translational error, with an RMSE of 1.030~m and a maximum APE of 2.426~m.These results show that FU-MPC provides substantially better robustness than EPIC in complex environments.

Overall, although FUEL and ERRT obtain low APE values in some cases, they do not successfully complete the exploration task due to local repetitive scanning.
The comparison with EPIC demonstrates that FU-MPC achieves lower translational APE while preserving task completion, indicating improved accuracy and robustness in the evaluated environments.

\subsection{Ablation Studies}



\begin{table}[H]
\centering
\caption{Translation APE and exploration efficiency comparison under different rotation strategies.}
\label{tab:rotation_ape_ablation}
\renewcommand{\arraystretch}{1.2}
\begin{tabular}{lcccc}
\hline\hline
Control Strategies & \makecell{Fixed \\30 deg/s} & \makecell{Fixed \\100 deg/s} & \makecell{Fixed \\360 deg/s} & FU-MPC \\
\hline
RMSE (m)        & 0.121 & 0.075 & 0.097 & \textbf{0.068} \\
Mean (m)        & 0.111 & \textbf{0.063} & 0.091 & \textbf{0.063} \\
Max (m)         & 0.351 & 0.569 & 0.176 & \textbf{0.161} \\
Trajectory (m)  & 309.68 & 276.76 & 237.92 & \textbf{214.94} \\
Exp.Time (s)    & 165.50 & 147.24 & 125.69 & \textbf{112.98} \\
\hline\hline
\end{tabular}
\end{table}

Table~\ref{tab:rotation_ape_ablation} reports the ablation results of different LiDAR rotation strategies in terms of translational APE.
Compared with fixed-speed rotation strategies, the proposed active rotation strategy achieves the lowest translational RMSE in the Spatial Maze map.
In the Spatial Maze map, the adaptive strategy reduces the RMSE to 0.068~m, outperforming the fixed 30~deg/s, 100~deg/s, and 360~deg/s settings, whose RMSE values are 0.121~m, 0.075~m, and 0.097~m, respectively.
It also obtains the lowest maximum APE of 0.161~m, indicating more stable pose estimation over the trajectory. Traditional constant speed strategies exhibit a severe accuracy-efficiency trade-off. Lower rates 30~deg/s incur excessive trajectory 309.68~m and time 165.5~s costs, whereas higher rates 360~deg/s degrade accuracy. FU-MPC successfully overcomes this bottleneck. By dynamically optimizing the control law, it completes the task with the minimal trajectory length 214.94~m and execution time 112.98~s, yielding reductions of 9.6\% and 10.1\%, respectively, compared to the Fixed 360~deg/s baseline.

Overall, the ablation study shows that actively regulating the LiDAR rotation speed can improve translational accuracy and enhance robustness in environments with different geometric characteristics.

\subsection{Real-World Evaluation}

\begin{figure}[t]
    \centering
    \includegraphics[width=\columnwidth]{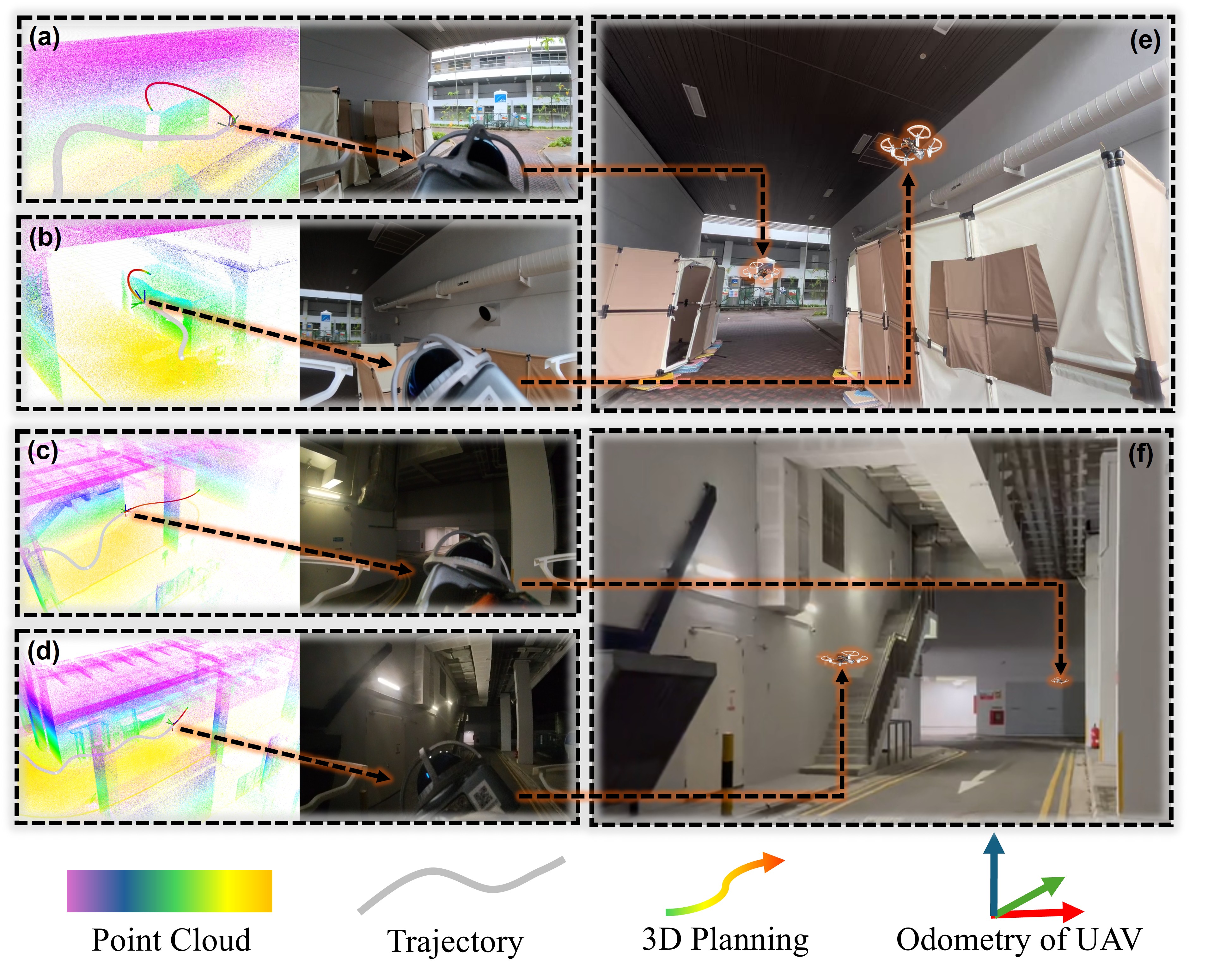}
    \caption{Real-world evaluation of the proposed FU-MPC in two different real scenarios. (a)-(d) The first-person perspective of UAV and the online mapping process. (e)-(f) The third-person perspective of UAV.}
    \label{fig:real_experiment}
\end{figure}

\begin{figure}[t]
    \centering
    \includegraphics[width=\columnwidth]{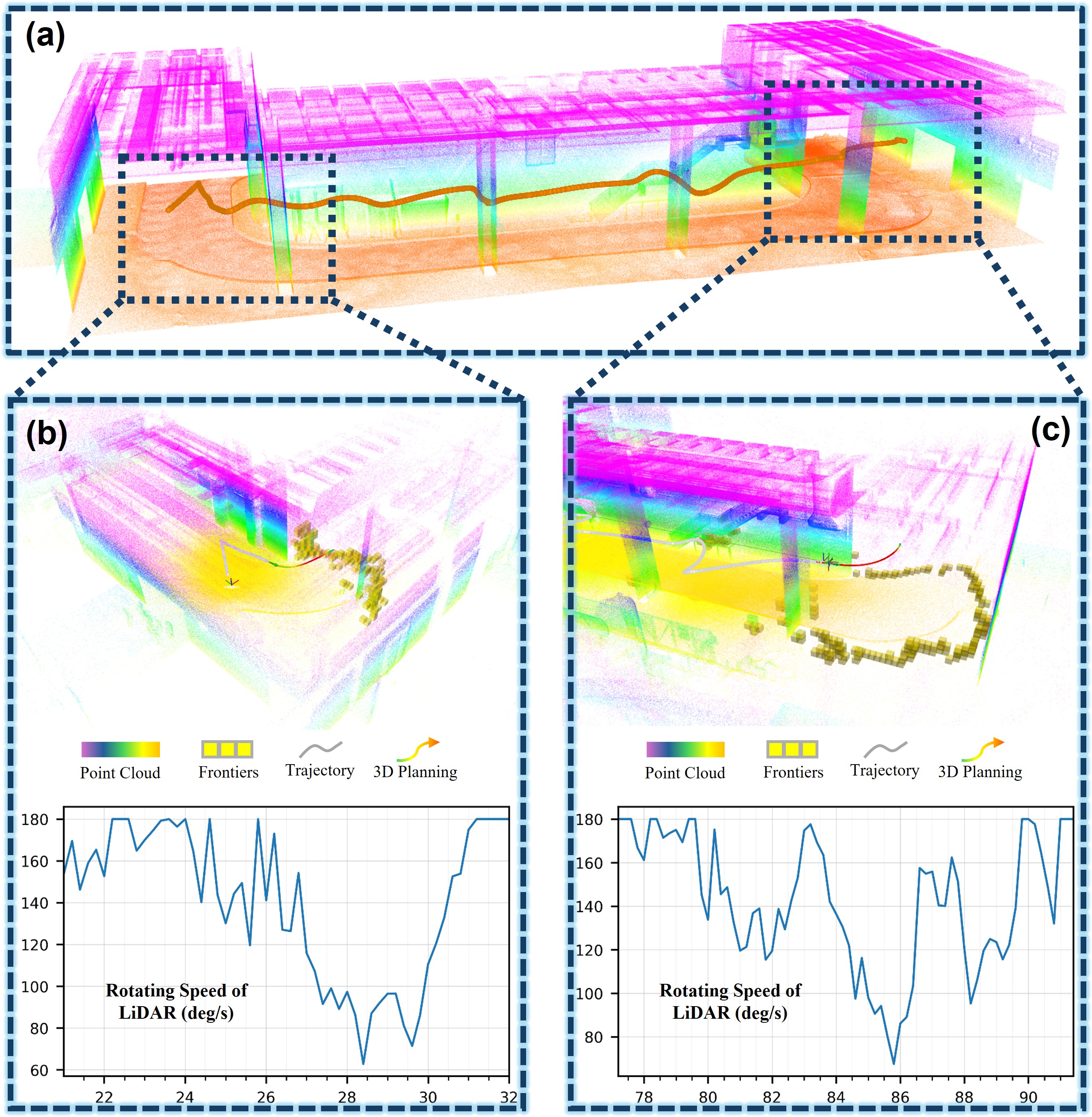}
    \caption{Real-world evaluation of the proposed FU-MPC. FU-MPC adaptively controls the LiDAR rotation speed via predictive scanning, improving both exploration efficiency and localization accuracy. (a) Overall trajectory in the industrial scene. (b)-(c) Predictive scanning at two representative corners and the rotating speed of LiDAR in two corners.}
    \label{fig:actvie_speed}
\end{figure}

To further validate the effectiveness of the proposed method, we conduct extensive field experiments in outdoor environments using the UAV platform shown in Fig.~\ref{fig:hardware}. The platform is equipped with an x86 Intel-N305 edge computing unit, enabling fully onboard computation. In all tests, the dynamics limits as $v_{max}$ = 0.5~m/s and $\omega_{max}$ =180~deg/s to ensure safety.

The first test scenario is a parking-lot tunnel containing two roofless house-like structures. The first-person and third-person views during the exploration process are presented in Fig.~\ref{fig:real_experiment}. In addition, Fig.~\ref{fig:real_experiment} illustrates the online mapping process, together with the historical trajectory and the planned trajectory of the UAV. In this scenario, the rotating LiDAR enables the UAV to scan the interior regions of the roofless structures from outside, without physically entering them. This capability improves both exploration efficiency and operational safety, particularly in confined or partially enclosed spaces.

We further evaluate the proposed method in a complex building-like environment, where the online mapping result is shown in Fig.~\ref{fig:actvie_speed}, and the corresponding exploration process is also presented in Fig.~\ref{fig:real_experiment}. This scenario contains two sharp corners, where the surrounding structure changes abruptly and poses challenges to both perception and localization. Benefiting from the proposed active LiDAR motion planning strategy, the UAV adaptively regulates the rotation speed of the LiDAR according to the local environmental complexity and motion prediction. As the UAV approaches a corner, the LiDAR rotation speed is reduced to improve scan overlap and localization accuracy through predictive scanning, as shown in Fig.~\ref{fig:actvie_speed}. After passing the corner and entering the corridor, the LiDAR rotation speed increases to enhance environmental coverage and exploration efficiency. More detailed results and visualizations of all real-world experiments are provided in the supplementary video.

\subsection{Real-Time Performance Analysis}

\begin{figure}[]
    \centering
    \includegraphics[width=\columnwidth]{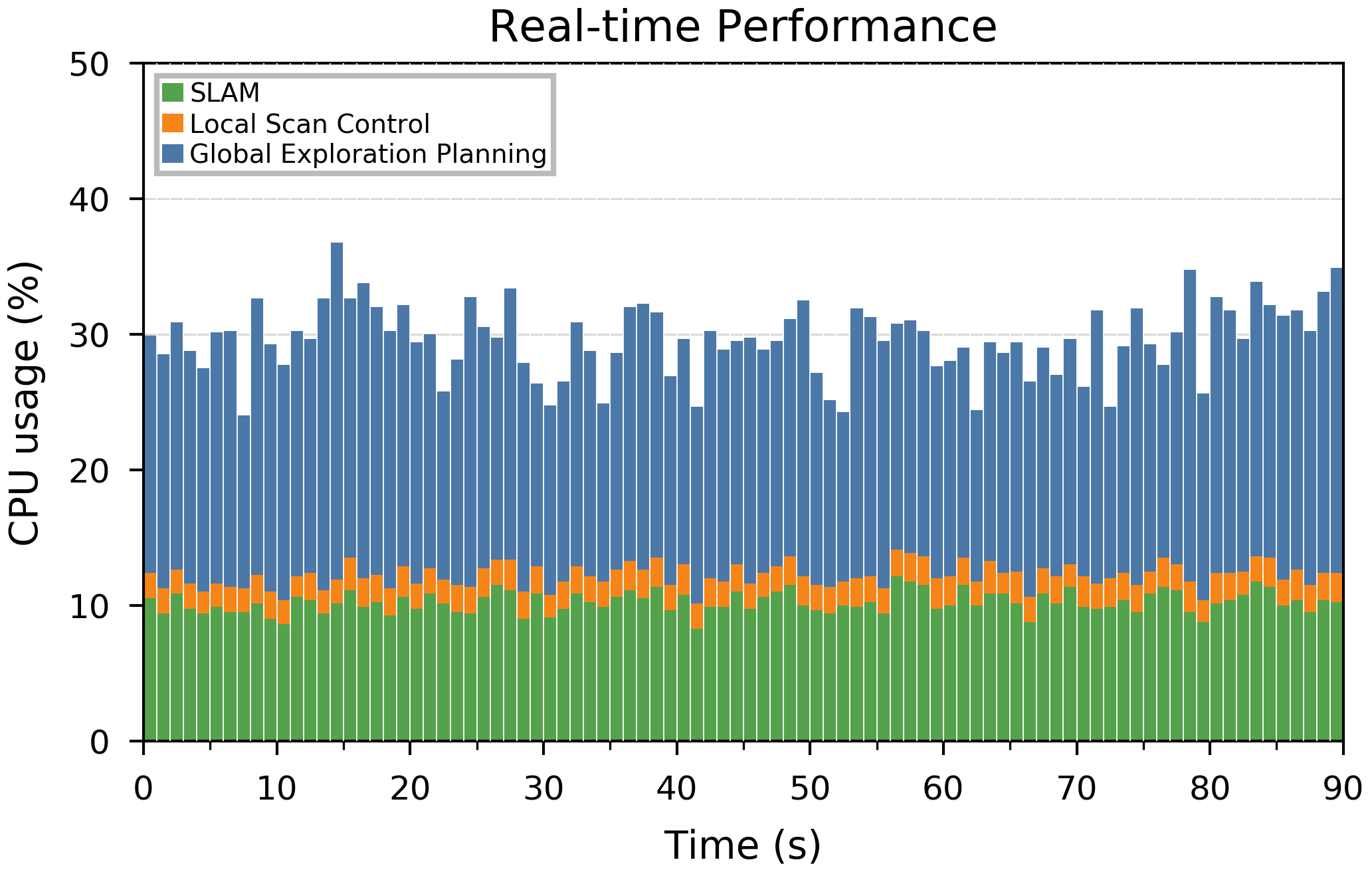}
    \caption{Real-time performance of the proposed control policy evaluated on an x86 Intel-N305 edge computing unit. Local Scan Control refers to the MPC controller of the rotating LiDAR. Global Exploration Planning includes global planning and local planning of UAV exploration.}
    \label{fig:real_time_performance}
\end{figure}

Beyond planning accuracy, the practical deployability of the proposed system also depends on its computational efficiency for real-time onboard execution. To this end, we further evaluate the runtime CPU utilization of the main onboard modules on an x86 Intel-N305 edge computing unit.

We benchmarked the CPU usage over a 90s real-world run, including the Global Exploration Planning module, Local Scan Control module, and SLAM module. As shown in Fig~\ref{fig:real_time_performance}. The Global Exploration Planning module consumes an average of 17.25\% CPU, with a peak of 24.88\%, while the Local Scan Control module remains lightweight, requiring only 1.88\% CPU on average and 2.38\% at peak. The SLAM module consumes 9.86\% CPU on average, with a peak of 12.13\%. Overall, the combined workload occupies 28.98\% CPU on average and remains below 36.75\% throughout the experiment.

These results demonstrate that the system satisfies real-time execution requirements while retaining substantial computational headroom for concurrent onboard tasks. In particular, the low overhead of our proposed method FU-MPC confirms that the proposed control module can be integrated with exploration planning and LiDAR-inertial SLAM without imposing a significant burden on the edge computing platform.

\section{Conclusion}

This paper presented FU-MPC, a Frontier- and Uncertainty-Aware Model Predictive Control framework for efficient and reliable UAV exploration with a motorized LiDAR. By treating LiDAR scanning as a controllable decision variable, we coupled two online-computable objectives: a predictive frontier gain derived from the future swept volume and a direction-dependent localization uncertainty metric constructed from local geometric constraints via Fisher information and an A-optimality criterion. To enable high-rate onboard execution, we introduced a lightweight piecewise-linear surrogate that significantly reduces the computational cost of utility evaluation and supports real-time receding-horizon optimization under hardware constraints. Experimental results demonstrate that FU-MPC improves exploration progress while maintaining robust SLAM performance compared with fixed-pattern scanning and uncertainty-only baselines.

\bibliographystyle{IEEEtran}
\bibliography{refs} 

@article{xu2022fast,
  title={Fast-lio2: Fast direct lidar-inertial odometry},
  author={Xu, Wei and Cai, Yixi and He, Dongjiao and Lin, Jiarong and Zhang, Fu},
  journal={IEEE Transactions on Robotics},
  volume={38},
  number={4},
  pages={2053--2073},
  year={2022},
  publisher={IEEE}
}

@inproceedings{zhang2014loam,
  title={LOAM: Lidar odometry and mapping in real-time.},
  author={Zhang, Ji and Singh, Sanjiv},
  booktitle={Robotics: Science and systems},
  volume={2},
  number={9},
  pages={1--9},
  year={2014},
  organization={Berkeley, CA}
}

@inproceedings{cui2024alphalidar,
  title={$\alpha$LiDAR: An Adaptive High-Resolution Panoramic LiDAR System},
  author={Cui, Jiahe and He, Yuze and Niu, Jianwei and Ouyang, Zhenchao and Xing, Guoliang},
  booktitle={Proceedings of the 30th Annual International Conference on Mobile Computing and Networking},
  pages={1515--1529},
  year={2024}
}

@article{li2025ua,
  title={Ua-mpc: Uncertainty-aware model predictive control for motorized lidar odometry},
  author={Li, Jianping and Xu, Xinhang and Liu, Jinxin and Cao, Kun and Yuan, Shenghai and Xie, Lihua},
  journal={IEEE Robotics and Automation Letters},
  year={2025},
  publisher={IEEE}
}

@article{geng2025epic,
  title={Epic: A lightweight lidar-based uav exploration framework for large-scale scenarios},
  author={Geng, Shuang and Ning, Zelin and Zhang, Fu and Zhou, Boyu},
  journal={IEEE Robotics and Automation Letters},
  year={2025},
  publisher={IEEE}
}

@article{alismail2015automatic,
  title={Automatic calibration of spinning actuated lidar internal parameters},
  author={Alismail, Hatem and Browning, Brett},
  journal={Journal of Field Robotics},
  volume={32},
  number={5},
  pages={723--747},
  year={2015},
  publisher={Wiley Online Library}
}

@article{kaul2016continuous,
  title={Continuous-time three-dimensional mapping for micro aerial vehicles with a passively actuated rotating laser scanner},
  author={Kaul, Lukas and Zlot, Robert and Bosse, Michael},
  journal={Journal of Field Robotics},
  volume={33},
  number={1},
  pages={103--132},
  year={2016},
  publisher={Wiley Online Library}
}

@article{shi2023real,
  title={Real-time multi-modal active vision for object detection on UAVs equipped with limited field of view LiDAR and camera},
  author={Shi, Chuanbeibei and Lai, Ganghua and Yu, Yushu and Bellone, Mauro and Lippiello, Vincezo},
  journal={IEEE Robotics and Automation Letters},
  volume={8},
  number={10},
  pages={6571--6578},
  year={2023},
  publisher={IEEE}
}

@inproceedings{bartolomei2021semantic,
  title={Semantic-aware active perception for uavs using deep reinforcement learning},
  author={Bartolomei, Luca and Teixeira, Lucas and Chli, Margarita},
  booktitle={2021 IEEE/RSJ International Conference on Intelligent Robots and Systems (IROS)},
  pages={3101--3108},
  year={2021},
  organization={IEEE}
}

@article{chen2024design,
  title={Design of an Adaptive Lightweight Lidar to decouple robot--camera geometry},
  author={Chen, Yuyang and Wang, Dingkang and Thomas, Lenworth and Dantu, Karthik and Koppal, Sanjeev J},
  journal={IEEE Transactions on Robotics},
  volume={40},
  pages={2254--2271},
  year={2024},
  publisher={IEEE}
}

@inproceedings{romero2024actor,
  title={Actor-critic model predictive control},
  author={Romero, Angel and Song, Yunlong and Scaramuzza, Davide},
  booktitle={2024 IEEE International Conference on Robotics and Automation (ICRA)},
  pages={14777--14784},
  year={2024},
  organization={IEEE}
}

@article{cao2023trust,
  title={Trust-region inverse reinforcement learning},
  author={Cao, Kun and Xie, Lihua},
  journal={IEEE Transactions on Automatic Control},
  volume={69},
  number={2},
  pages={1037--1044},
  year={2023},
  publisher={IEEE}
}

@article{chen2023self,
  title={A self-rotating, single-actuated UAV with extended sensor field of view for autonomous navigation},
  author={Chen, Nan and Kong, Fanze and Xu, Wei and Cai, Yixi and Li, Haotian and He, Dongjiao and Qin, Youming and Zhang, Fu},
  journal={Science Robotics},
  volume={8},
  number={76},
  pages={eade4538},
  year={2023},
  publisher={American Association for the Advancement of Science}
}

@inproceedings{feng2024fc,
  title={Fc-planner: A skeleton-guided planning framework for fast aerial coverage of complex 3d scenes},
  author={Feng, Chen and Li, Haojia and Zhang, Mingjie and Chen, Xinyi and Zhou, Boyu and Shen, Shaojie},
  booktitle={2024 IEEE International Conference on Robotics and Automation (ICRA)},
  pages={8686--8692},
  year={2024},
  organization={IEEE}
}

@article{yan2021sampling,
  title={Sampling-based path planning for high-quality aerial 3D reconstruction of urban scenes},
  author={Yan, Feihu and Xia, Enyong and Li, Zhaoxin and Zhou, Zhong},
  journal={Remote Sensing},
  volume={13},
  number={5},
  pages={989},
  year={2021},
  publisher={MDPI}
}

@article{zhou2020offsite,
  title={Offsite aerial path planning for efficient urban scene reconstruction},
  author={Zhou, Xiaohui and Xie, Ke and Huang, Kai and Liu, Yilin and Zhou, Yang and Gong, Minglun and Huang, Hui},
  journal={ACM Transactions on Graphics (TOG)},
  volume={39},
  number={6},
  pages={1--16},
  year={2020},
  publisher={ACM New York, NY, USA}
}

@article{feng2023predrecon,
  title={Predrecon: A prediction-boosted planning framework for fast and high-quality autonomous aerial reconstruction},
  author={Feng, Chen and Li, Haojia and Gao, Fei and Zhou, Boyu and Shen, Shaojie},
  journal={arXiv preprint arXiv:2302.04488},
  year={2023}
}

@inproceedings{yamauchi1997frontier,
  title={A frontier-based approach for autonomous exploration},
  author={Yamauchi, Brian},
  booktitle={Proceedings 1997 IEEE International Symposium on Computational Intelligence in Robotics and Automation CIRA'97.'Towards New Computational Principles for Robotics and Automation'},
  pages={146--151},
  year={1997},
  organization={IEEE}
}

@article{zhou2021fuel,
  title={Fuel: Fast uav exploration using incremental frontier structure and hierarchical planning},
  author={Zhou, Boyu and Zhang, Yichen and Chen, Xinyi and Shen, Shaojie},
  journal={IEEE Robotics and Automation Letters},
  volume={6},
  number={2},
  pages={779--786},
  year={2021},
  publisher={IEEE}
}

@article{zhou2023racer,
  title={Racer: Rapid collaborative exploration with a decentralized multi-uav system},
  author={Zhou, Boyu and Xu, Hao and Shen, Shaojie},
  journal={IEEE Transactions on Robotics},
  volume={39},
  number={3},
  pages={1816--1835},
  year={2023},
  publisher={IEEE}
}

@inproceedings{yamauchi1998frontier,
  title={Frontier-based exploration using multiple robots},
  author={Yamauchi, Brian},
  booktitle={Proceedings of the second international conference on Autonomous agents},
  pages={47--53},
  year={1998}
}

@inproceedings{cao2021tare,
  title={TARE: A hierarchical framework for efficiently exploring complex 3D environments.},
  author={Cao, Chao and Zhu, Hongbiao and Choset, Howie and Zhang, Ji},
  booktitle={Robotics: Science and Systems},
  volume={5},
  pages={2},
  year={2021}
}

@inproceedings{tang2023bubble,
  title={Bubble explorer: Fast UAV exploration in large-scale and cluttered 3D-environments using occlusion-free spheres},
  author={Tang, Benxu and Ren, Yunfan and Zhu, Fangcheng and He, Rui and Liang, Siqi and Kong, Fanze and Zhang, Fu},
  booktitle={2023 IEEE/RSJ International Conference on Intelligent Robots and Systems (IROS)},
  pages={1118--1125},
  year={2023},
  organization={IEEE}
}

@inproceedings{bircher2016receding,
  title={Receding horizon" next-best-view" planner for 3d exploration},
  author={Bircher, Andreas and Kamel, Mina and Alexis, Kostas and Oleynikova, Helen and Siegwart, Roland},
  booktitle={2016 IEEE international conference on robotics and automation (ICRA)},
  pages={1462--1468},
  year={2016},
  organization={IEEE}
}

@article{lindqvist2024tree,
  title={A tree-based next-best-trajectory method for 3-D UAV exploration},
  author={Lindqvist, Bj{\"o}rn and Patel, Akash and L{\"o}fgren, Kalle and Nikolakopoulos, George},
  journal={IEEE Transactions on Robotics},
  volume={40},
  pages={3496--3513},
  year={2024},
  publisher={IEEE}
}

@inproceedings{cieslewski2017rapid,
  title={Rapid exploration with multi-rotors: A frontier selection method for high speed flight},
  author={Cieslewski, Titus and Kaufmann, Elia and Scaramuzza, Davide},
  booktitle={2017 IEEE/RSJ International Conference on Intelligent Robots and Systems (IROS)},
  pages={2135--2142},
  year={2017},
  organization={IEEE}
}

@inproceedings{xu2024vrexplorer,
  title={VRExplorer: An efficient view-region based autonomous exploration method in unknown environments for UAV},
  author={Xu, Kai and Zheng, Lanxiang and Wei, Mingxin and Cheng, Hui},
  booktitle={2024 IEEE/RSJ International Conference on Intelligent Robots and Systems (IROS)},
  pages={8081--8087},
  year={2024},
  organization={IEEE}
}

@article{huang2010observability,
  title={Observability-based rules for designing consistent EKF SLAM estimators},
  author={Huang, Guoquan P and Mourikis, Anastasios I and Roumeliotis, Stergios I},
  journal={The international journal of Robotics Research},
  volume={29},
  number={5},
  pages={502--528},
  year={2010},
  publisher={SAGE Publications Sage UK: London, England}
}

@inproceedings{burgard2005information,
  title={Information gain-based exploration using rao-blackwellized particle filters},
  author={Burgard, Wolfram},
  booktitle={Robotics: science and systems I},
  year={2005}
}

@inproceedings{koga2022active,
  title={Active SLAM over continuous trajectory and control: A covariance-feedback approach},
  author={Koga, Shumon and Asgharivaskasi, Arash and Atanasov, Nikolay},
  booktitle={2022 American Control Conference (ACC)},
  pages={5062--5068},
  year={2022},
  organization={IEEE}
}

@article{faria2019autonomous,
  title={Autonomous 3D exploration of large structures using an UAV equipped with a 2D LIDAR},
  author={Faria, Margarida and Ferreira, Ant{\'o}nio S{\'e}rgio and P{\'e}rez-Leon, H{\'e}ctor and Maza, Ivan and Viguria, Antidio},
  journal={Sensors},
  volume={19},
  number={22},
  pages={4849},
  year={2019},
  publisher={MDPI}
}

@article{zhang2023unmanned,
  title={Unmanned aerial vehicle navigation in underground structure inspection: A review},
  author={Zhang, Ran and Hao, Guangbo and Zhang, Kong and Li, Zili},
  journal={Geological Journal},
  volume={58},
  number={6},
  pages={2454--2472},
  year={2023},
  publisher={Wiley Online Library}
}

@article{kerle2019uav,
  title={UAV-based structural damage mapping: A review},
  author={Kerle, Norman and Nex, Francesco and Gerke, Markus and Duarte, Diogo and Vetrivel, Anand},
  journal={ISPRS international journal of geo-information},
  volume={9},
  number={1},
  pages={14},
  year={2019},
  publisher={MDPI}
}

@article{placed2023survey,
  title={A survey on active simultaneous localization and mapping: State of the art and new frontiers},
  author={Placed, Julio A and Strader, Jared and Carrillo, Henry and Atanasov, Nikolay and Indelman, Vadim and Carlone, Luca and Castellanos, Jos{\'e} A},
  journal={IEEE Transactions on Robotics},
  volume={39},
  number={3},
  pages={1686--1705},
  year={2023},
  publisher={IEEE}
}

@article{xu2021fast,
  title={Fast-lio: A fast, robust lidar-inertial odometry package by tightly-coupled iterated kalman filter},
  author={Xu, Wei and Zhang, Fu},
  journal={IEEE Robotics and Automation Letters},
  volume={6},
  number={2},
  pages={3317--3324},
  year={2021},
  publisher={IEEE}
}

@inproceedings{zhang2019beyond,
  title={Beyond point clouds: Fisher information field for active visual localization},
  author={Zhang, Zichao and Scaramuzza, Davide},
  booktitle={2019 International Conference on Robotics and Automation (ICRA)},
  pages={5986--5992},
  year={2019},
  organization={IEEE}
}

@article{chen2022direct,
  title={Direct lidar-inertial odometry: Lightweight lio with continuous-time motion correction},
  author={Chen, Kenny and Nemiroff, Ryan and Lopez, Brett T},
  journal={arXiv preprint arXiv:2203.03749},
  year={2022}
}

@article{chen2022r,
  title={R-LIO: Rotating LiDAR inertial odometry and mapping},
  author={Chen, Kai and Zhan, Kai and Pang, Fan and Yang, Xiaocong and Zhang, Da},
  journal={Sustainability},
  volume={14},
  number={17},
  pages={10833},
  year={2022},
  publisher={MDPI}
}

@article{placed2022general,
  title={A general relationship between optimality criteria and connectivity indices for active graph-SLAM},
  author={Placed, Julio A and Castellanos, Jos{\'e} A},
  journal={IEEE Robotics and Automation Letters},
  volume={8},
  number={2},
  pages={816--823},
  year={2022},
  publisher={IEEE}
}

@article{zhu2025flare,
  title={FLARE: Fast Autonomous Aerial Exploration in Large-Scale 3D Scenarios Using Actively Rotated LiDAR},
  author={Zhu, Zhiwen and Fang, Yuhao and Xiao, Xulin and Lyu, Ximin and Mei, Jie and Zhou, Boyu},
  journal={IEEE Transactions on Automation Science and Engineering},
  volume={22},
  pages={24077--24091},
  year={2025},
  publisher={IEEE}
}

\raggedbottom
\vspace{-1.5cm}
\begin{IEEEbiographynophoto}
{Jianping Li (Member, IEEE)}
received the B.S. degree in geographic information system (GIS) and the Ph.D. degree in photogrammetry and remote sensing from Wuhan University, Wuhan, China, in 2015 and 2021, respectively. He is currently a Research Fellow with the School of Electrical and Electronic Engineering, Nanyang Technological University, Singapore. His research interests include 3-D sensing system integration, uncrewed aerial vehicle (UAV)/uncrewed ground vehicle (UGV) mapping, robot perception, and point cloud data processing, with over 50 publications in venues, such as ISRPS, IEEE TITS, IEEE TGRS, RAL, ICRA, and IROS. He received the Science and Technology Progress Award in Surveying and Mapping in 2019 and 2023.
\end{IEEEbiographynophoto}
\vspace{-1.5cm}

\begin{IEEEbiographynophoto}
{Pengfei Wan}
received the B.Eng. degree in automation from Northwestern Polytechnical University, Xi’an, China, in 2022. He is currently pursuing the M.S. degree in computer control and automation with the School of Electrical and Electronic Engineering, Nanyang Technological University, Singapore. His current research interests include uncrewed aerial vehicle (UAV) mapping, autonomous exploration, and motion planning.
\end{IEEEbiographynophoto}

\vspace{-1.5cm}
\begin{IEEEbiographynophoto}
{Zhongyuan Liu}
received the B.Sc. (Hons.) degree in telecommunications engineering with management from Queen Mary University of London, U.K., and Beijing University of Posts and Telecommunications, China, in 2024. He is the Co-Founder of CertaintyX, Singapore. His research interests include sensor calibrationx and fusion, active SLAM, and robotic navigation.
\end{IEEEbiographynophoto}

\vspace{-1.5cm}
\begin{IEEEbiographynophoto}
{Yi Wang}
received the B.E. degree in robotics engineering from Zhejiang University, Hangzhou, China, in 2024, and the M.Eng. degree in control science and engineering from Zhejiang University, Hangzhou, China, and the M.Sc. degree in Electrical and Electronic Engineering from Nanyang Technological University, Singapore, in 2026. His research interests include aerial manipulation, and motion planning.
\end{IEEEbiographynophoto}

\vspace{-1.5cm}
\begin{IEEEbiographynophoto}
{Yiheng Chen}
is currently pursuing the B.Eng. degree in mechanical engineering with The Hong Kong Polytechnic University, Hong Kong. He was an exchange student at the National University of Singapore. His current research interests include autonomous aerial vehicles, autonomous exploration, robotic perception, motion planning, and intelligent control systems.
\end{IEEEbiographynophoto}

\vspace{-1.5cm}
\begin{IEEEbiographynophoto}
{Xinhang Xu} 
received the B.Eng. degree in automation from the South China University of Technology, Guangzhou, China, and the M.Sc. degree in Electrical and Electronic Engineering from Nanyang Technological University, Singapore, in 2023. He is currently pursuing the Ph.D. degree in Electrical and Electronic Engineering at Nanyang Technological University, Singapore. His research interests include intelligent sensing, social navigation, aerial manipulation, and tethered robotics.
\end{IEEEbiographynophoto}

\vspace{-1.5cm}
\begin{IEEEbiographynophoto}
{Rui Jin}
received the B.E. degree in mechanical design, manufacturing and automation from Northwestern Polytechnical University, Xi'an, China, in 2021, and the M.Eng. degree in control science and engineering from Zhejiang University, Hangzhou, China, in 2024. He is currently working toward the Ph.D. degree with the School of Electrical and Electronic Engineering, Nanyang Technological University, Singapore, under the supervision of Prof. Lihua Xie. His research interests include aerial manipulation, and motion planning.
\end{IEEEbiographynophoto}

\vspace{-1.5cm}
\begin{IEEEbiographynophoto}
{Boyu Zhou (Member, IEEE)} received the Ph.D. degree in electronic and computer engineering from The Hong Kong University of Science and Technology, Hong Kong, China, in 2022. He is currently
a tenure-track Assistant Professor (doctoral supervisor) with the Southern University of Science and Technology (SUSTech), where he is also the Director of the Smart Autonomous Robotics (STAR) Group. His research interests encompass aerial and mobile robots, motion planning, active perception, multi-robot systems, mobile manipulation, and vision-language navigation. He was a recipient of the IEEE TRO 2023 King-Sun Fu Best Paper Award, the IEEE RAL 2023 Best Paper Award, and the IEEE ICRA 2024 Best UAV Paper Finalist. He is an Associate Editor of IEEE TRANSACTIONS ON ROBOTICS and IEEE ICRA.
\end{IEEEbiographynophoto}

\vspace{-1.5cm}
\begin{IEEEbiographynophoto}
{Lihua Xie (Fellow, IEEE)}
received the Ph.D. degree in electrical engineering from the University of Newcastle, Callaghan, NSW, Australia, in 1992. He was the Head of the Division of Control and Instrumentation and the Co-Director of the Delta-NTU Corporate Laboratory for Cyber-Physical Systems. He is currently a President’s Chair Professor and the Director of the Center for Advanced Robotics Technology Innovation, Nanyang Technological University, Singapore. His research interests include robust control and estimation, networked control systems, multiagent systems, and uncrewed systems. Dr. Xie is a Fellow of the Academy of Engineering Singapore, IFAC, and CAA. He is currently the Editor-in-Chief of \textit{Unmanned Systems}. He has served as an Editor for the IET Book Series in Control and as an Associate Editor for several journals, including \textit{IEEE Transactions on Automatic Control}, \textit{Automatica}, \textit{IEEE Transactions on Control Systems Technology}, \textit{IEEE Transactions on Control of Network Systems}, and \textit{IEEE Transactions on Circuits and Systems II: Express Briefs}. He was an IEEE Distinguished Lecturer from 2012 to 2014.
\end{IEEEbiographynophoto}

\end{document}